\RequirePackage[loading]{tracefnt}

\documentclass[letterpaper, 10 pt, conference]{ieeeconf}
\renewcommand{\baselinestretch}{.98} 
\IEEEoverridecommandlockouts                              %

\usepackage{times}
\usepackage{xintexpr,colortbl}
\usepackage[keeplastbox]{flushend}

\usepackage{multicol}
\usepackage{hyperref}

\usepackage{graphicx}
\usepackage{tabularx} %
\usepackage{multirow,diagbox}
\usepackage{adjustbox}
\usepackage{amsmath,amsfonts,amssymb}
\usepackage{gensymb}
\usepackage{xcolor}
\usepackage{tikz}

\definecolor{amaranth}{rgb}{0.9, 0.17, 0.31}
\definecolor{redryb}{rgb}{1.0, 0.17, 0.07}
\newcommand*{\MinNumber}{0}%
\newcommand*{\MaxNumber}{3.5}%

\newcommand{\ApG}[1]{%
  \pgfmathsetmacro{\PercentColor}{abs(100.0*(#1-\MinNumber)/(\MaxNumber-\MinNumber))}%
  \edef\x{\noexpand\cellcolor{redryb!\PercentColor}}\x\textcolor{black}{#1}%
}

\usepackage{subcaption}

\title{\LARGE \bf
SOIC: Semantic Online Initialization and Calibration \\ for LiDAR and Camera}

\author{Weimin Wang, Shohei Nobuhara, Ryosuke Nakamura and Ken Sakurada
\thanks{This work was supported in part by the New Energy and Industrial Technology Development Organization (NEDO) and in part by an internal grant of National Institute of Advanced Industrial Science and Technology (AIST).}%
\thanks{W. Wang, R. Nakamura and K. Sakurada are with National Institute of Advanced Industrial Science and Technology (AIST), Tokyo, 135-0064, Japan
        {\tt\small (email: weimin.wang@aist.go.jp; r.nakamura@aist.go.jp; k.sakurada@aist.go.jp).}}%
\thanks{S. Nobuhara is with the Department of Intelligence Science and Technology, Graduate School of Informatics, Kyoto University, Kyoto, 606-8501, Japan
        {\tt\small (email: nob@i.kyoto-u.ac.jp).}}%
}

\begin{document}

\maketitle
\thispagestyle{empty}
\pagestyle{empty}

\begin{abstract}
This paper presents a novel  semantic-based online extrinsic calibration approach, SOIC (so, I see), for Light Detection and Ranging (LiDAR) and camera sensors. Previous online calibration methods usually need prior knowledge of rough initial values for optimization. The proposed approach removes this limitation by converting the initialization problem to a Perspective-n-Point (PnP) problem with the introduction of semantic centroids (SCs). The closed-form solution of this PnP problem has been well researched and can be found with existing PnP methods. Since the semantic centroid of the point cloud usually does not accurately match with that of the corresponding image, the accuracy of parameters are not improved even after a nonlinear refinement process. Thus, a cost function based on the constraint of the correspondence between semantic elements from both point cloud and image data is formulated. Subsequently, optimal extrinsic parameters are estimated by minimizing the cost function. We evaluate the proposed method either with GT or predicted semantics on KITTI dataset. Experimental results and comparisons with the baseline method verify the feasibility of the initialization strategy and the accuracy of the calibration approach. In addition, we release the source code at \href{https://github.com/}{https://github.com/$--$/SOIC}.

\end{abstract}

\IEEEpeerreviewmaketitle

\section{INTRODUCTION}
Light Detection and Ranging (LiDAR) sensors are able to obtain spatial data robustly in a wide range but with low resolution and no color, while camera sensors obtain RGB image at a high resolution but light-sensitive and no distance information. To compensate for the weakness of each other, the combination of LiDAR and camera sensors have been a typical and essential setup for applications in mobile robotics and autonomous driving vehicles.
Based on this combination, neural networks such as MV3D \cite{Chen2016Multiview3O}, AVOD \cite{Ku2017Joint3P} and F-PonitNet\cite{Qi2017FrustumPF} are proposed to improve the performance for traditional object detection and segmentation tasks.  As the most-critical precondition for the combination, an accurate extrinsic calibration, which estimates the transformation matrix between coordinate systems of the two sensors, is usually a first and vital step.

\begin{figure}[t]
\centering
\includegraphics[width=0.47\textwidth]{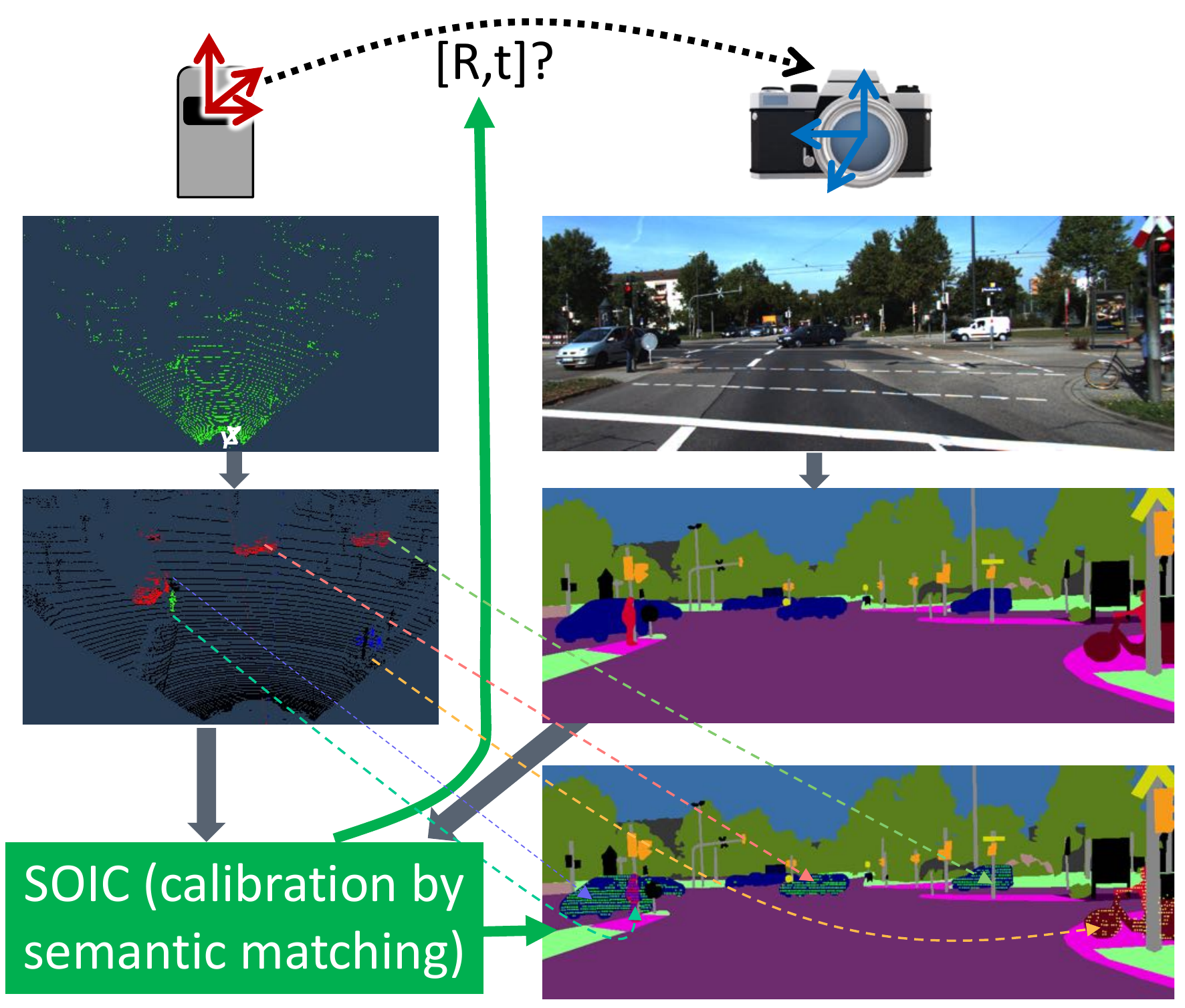}
  \caption{SOIC estimates the extrinsic calibration parameters between LiDAR and camera sensors based on the semantic matching of point cloud and image data.}
  \label{fig:intro}
\end{figure}

Many LiDAR-camera calibration methods have been proposed in the past years. Traditionally, manual interventions are usually needed to either select features or correspondences between features from point clouds and images \cite{unnikrishnan_hebert_2018,Moghadam2013LinebasedEC,Park_2014}. To improve the convenience of the process, methods that can automatically correspond to the detected features are proposed \cite{Geiger_2012,Wang2017ReflectanceIA}. Specific targets like chessboard are necessary for these methods. To increase the flexibility, online target-less methods are proposed. One approach is based on observations that find extrinsic parameters by utilizing the correlation of intensity or edges between the observed point cloud and image data \cite{Levinson2013AutomaticOC,MIcali,GOMcali,GMMcali}. A learning-based method is proposed to learn this correlation by neural networks \cite{Iyer2018CalibNetGS}.
The performance of these methods usually greatly depends on the accuracy of the initial guess. 
Another target-less approach is to obtain the calibration parameters by matching motions of the two sensors \cite{taylor2015motion,Park2020SpatiotemporalCC}. Sufficient and accurate ego-motion estimations are required to achieve high accuracy. 

In this paper we propose the \textbf{semantic} calibration method, Semantic Online and Initialization Calibration (SOIC), to address the initialization challenge, by utilizing the semantic segmentation results of point cloud and image data. 
As demonstrated in Fig.~\ref{fig:intro}, SOIC estimates the initial guess and the final calibration parameters with the semantic segmentation results which are usually obtainable in the perceptions for intelligent robots or autonomous driving cars.
Since SOIC works as so long as there are enough variety of semantic object changes instead of the whole scene, it can even be used for calibration for offline scenarios (e.g., indoor robotics). More specifically, our contributions can be concluded as follows:
\begin{itemize}
\item Propose a novel online target-less approach for LiDAR-camera extrinsic calibration based on semantic segmentation which is usually an inevitable process for most AI agent applications.
\item Introduce the semantic centroid (SC) to estimate the initial values for the optimization. To the best of our knowledge, this is the first target-less online calibration method with no need for prior initial values.
\item Evaluate the proposed method on KITTI dataset to validate the feasibility. Moreover, we apply SOIC with prediction results by existing networks to confirm the practicality.
\item Make the source code publicly available.
\end{itemize}

\section{RELATED WORK}

For observation-based approaches, the essence of extrinsic calibration of LiDAR-camera sensors can be considered as three processes: 1) Extracting 2D-3D features that can be observed by both sensors, such as corners and edges. 2) Corresponding the extracted features. This process may need user intervention depending on the method.  3) Optimizing a defined cost function to minimize the distance of corresponding features at a defined metric. We revisit related work from various features and subsequent different approaches with different features.

{\bf Handcrafted features}. 
In spite of the fact that there exist many methods that extract handcrafted features like SIFT \cite{Lowe2004DistinctiveIF} on 2D images and 3D-SIFT \cite{3DSIFT} on the point cloud, it is still challenging to find the common features and establish the correspondence due to the cross-domain gap of 2D and 3D data. Recently, learning-based methods of descriptors for handcrafted or learned 2D (e.g., SIFT \cite{Lowe2004DistinctiveIF}) and 3D features (e.g., ISS \cite{Zhong2009IntrinsicSS}) to bridge the correspondence of two different domains are also proposed \cite{Feng20192D3DMatchnetLT,pham2020lcd}. Networks are trained to extract descriptors of the range of interest (ROI) area, which usually is a path of an image or nearest neighboring points within a certain radius of a point cloud of detected feature pixels and points.  

{\bf Target-based methods.} 
Traditional methods usually achieve the extrinsic calibration by utilizing specific target objects owning common features that can be detected and corresponded in images and point clouds acquired by the two sensors. These possible features can be planarity \cite{Mirzaei20123DLI}, edges of artificial shapes of a board \cite{Velas2014CalibrationOR, Guindel2017AutomaticEC}, corners of monochromatic board \cite{Park2014CalibrationBC}, pattern corners of chessboard \cite{Wang2017ReflectanceIA}. 
Given such correspondences, target-based methods are able to give an accurate estimation of extrinsic parameters. However, preparation and processing usually take time and effort. Moreover, they are less feasible in online situations due to the necessity of target objects. 

{\bf Target-less methods.} 
In applications like autonomous driving, a calibrated LiDAR-camera set may be drifted or even collapsed due to vibration and harshness in driving. This may lead to serious safety problems. To loose the restriction of artificial targets or manual operations and make extrinsic calibration more flexible, target-less methods are also proposed \cite{Levinson2013AutomaticOC,MIcali,GOMcali,GMMcali}. These methods usually find optimal parameters by maximizing the correlation of general features distributions like intensity or edges extracted from two modalities data. The generality of these features makes it possible to get rid of target objects. Nevertheless, it increases the possibilities of local optima due to the ambiguity caused by this generality. That is a reason why existing target-less methods suppose a rough initial guess is available.

The difference between target-based and target-less calibration methods can be regarded as the bias of ``uniqueness'' and ``targetlessness'' of features. High ``uniqueness'' brings advantages for matching and optimization, while ``targetlessness'' lowers the dependencies on the calibration environment. For example, distinguishable pattern corners from chessboards in target-based methods can be used for orientated matching owing to their high ``uniqueness'', while general features in the target-less methods are just the opposite. Thus, we propose the semantic-based approach to take the trade-off between these two aspects.

\section{METHOD}
The proposed method consists of three steps. First, we apply existing methods to get semantic segmentation results on images and point cloud with pre-trained models. With these segmented results from multiple pairs of image and point cloud frames, we estimate an initial coarse pose based on the semantic centroids (SCs). A cost function formulated under the constraints of the semantic correspondence is further defined. Finally, the fine parameters are discovered by optimizing the defined cost function with coarse pose as the initial guess. Details of the proposed method are presented in the following part.

\subsection{Problem Formulation}
For a pair of point cloud and image, we notate
the point cloud as $P^L=\{ \textbf{p}_{1}^L,\textbf{p}_{2}^L,\ldots ,\textbf{p}_{n}^L\}$, where $\textbf{p}_{i}^L=(x_i,y_i,z_i) \in \mathbb{R}^3$ and $n$ is the number of points in the point cloud. The superscript $L$ indicates the LiDAR coordinate system.

In addition,  we have $\ell_i^{pcd} \in S$ for point $p_i$ indicating the semantic label, where $S=\{0,1,2...N\}$ means the set of semantic classes. 

Similarly, for the image $I$ with the width $W$ and height $H$, we also have $\ell_{[l,m]}^{img} \in S$ indicating the label of pixel $I{[l,m]}$, where $l\in [0,W]$ and $m\in [0,H]$. Due to the resolution difference, the number of pixels is much more than the number of point cloud.

We define the problem as finding a rotation angle vector $\boldsymbol{\theta}=(\theta_x,\theta_y,\theta_z)$
and translation vector
$\textbf{t}=(t_x,t_y,t_z)$ to transform $P^L$ to $P^C$ such that as many points are projected to pixels of image $I$ with the same semantic class. The superscript $C$ of $P^C$ indicates the camera coordinate system.

A point $\textbf{p}_i^L$ in the LiDAR coordinate system can be transformed to the camera coordinate system with $\boldsymbol{\theta}$ and $\textbf{t}$ as:
\begin{equation}
\begin{split}
        \textbf{p}_i^C =\boldsymbol{\mathcal{R}}(\boldsymbol{\theta})\cdot \textbf{p}_i^L+\textbf{t}.
\end{split}
\label{eq:trans}
\end{equation}

If the intrinsic parameters $\textbf{K}$ and the projection function $\boldsymbol{\mathcal{P}}$ of the camera are known, we can project the 3D point $\textbf{p}_{i}^C$ to the pixel coordinates {$[u^i,v^i]$} of image by:
\begin{equation}
    [u^i,v^i]=\boldsymbol{\mathcal{P}}(\textbf{K},\textbf{p}_{i}^C).
    \label{eq:cam_proj}
\end{equation}
Note that the point maybe projected out of the image range due to bad extrinsic parameters. It means that $[u^i,v^i]$ may exceed the range of $[0,W]$ and $[0,H]$ respectively.

As previously stated, the constraints of the cost function should be designed to maximize the consistency between the label $\ell_i^{pcd} \in S$ of the point cloud $p_i$ and the label $\ell_{[u^i,v^i]}^{img} \in S$ of the project pixel $[u^i,v^i]$. We define the the consistency function $\boldsymbol{\mathcal{C}}$ as:
\begin{equation}
    \boldsymbol{\mathcal{C}} = 1 - e^{-\epsilon^{-1}\left |  (\ell_i^{pcd}-\ell_{[u^i,v^i]}^{img}))\right |},
    \label{eq:cons}
\end{equation}
where $\epsilon$ represents a very small number such that $e^{-\epsilon^{-1}}$ approaches to zero. Thus, if the labels of $p_i$ and $[u^i,v^i]$ are consistent, $\mathcal{C}$ will be $0$ \textit{i.e.}, no cost for $p_i$. Inversely,  $\boldsymbol{\mathcal{C}}$ approximates to $1$ to account the cost for $p_i$ if the labels are inconsistent. Although no derivative is necessary for the optimization method we used in this work, the differentiability of Eq.~\ref{eq:cons} is useful for methods like Gradient Descent which needs analytical derivative calculation.

For the transformed point $p_i^C$ that is projected out of the image or projected to a pixel with inconsistent semantic label, the cost for the original point $p_i^L$ is calculated with a defined distance function $\boldsymbol{\mathcal{D}}$ defined as:
\begin{equation}
    \boldsymbol{\mathcal{D}}(\textbf{p}_{i}^L)=\min_{\substack{\ell^{img}_{[l,m]} = \ell^{pcd}_i  \\ l\in [0,W], m\in [0,H] }}(\boldsymbol{\mathcal{M}}([u,v]^i,[l,m]))\left|{p}_{i}^L \right |^2.
    \label{eq:dis_cost}
\end{equation}
Basically, this function calculates the minimum Manhattan distance of the pixel in the image $I$ with the same label of $p_i^L$.
In Eq.~\ref{eq:dis_cost}, $\boldsymbol{\mathcal{M}}$ denotes the Manhattan distance defined as:
\begin{equation}
    \boldsymbol{\mathcal{M}}([u,v],[l,m])=\left |u-l\right |+\left |v-m\right |.
    \label{eq:manhattan}
\end{equation}

There usually exist more than one semantic class. Thus, it is better to perform the semantic alignment and cost calculation ``classwisely'' to weight the cost class-by-class. Accordingly, we utilize the indicator function ${\mathbf {1} _{A}}$ defined as:
\begin{equation}
\begin{split}
    {\mathbf {1} _{A}(x):={\begin{cases}1&{\text{if }}x\in A,\\0&{\text{if }}x\notin A.\end{cases}}}
\end{split}
\label{equ:iniciator}
\end{equation}

Combining all the functions described from Eqs.~\ref{eq:trans}-\ref{equ:iniciator}, we get the final cost function $\mathcal{L}$ for one pair of point cloud and image defined in Eq.~\ref{eq:cost_fun}. The denominator indicates the number of points with valid semantic labels as the normalization for multiple pairs. 
\begin{equation}
    {\mathcal{L}} =\frac{\sum_{s\in S} \sum_{i}^{n}\mathbf {1} _{\{s\}}(\ell_i^{pcd})\boldsymbol{\mathcal{C}}(\textbf{p}_{i}^L)\boldsymbol{\mathcal{D}}(\textbf{p}_{i}^L)}{\sum_{s\in S} \sum_{i}^{n}\mathbf {1} _{\{s\}}(\ell_i^{pcd})}
    \label{eq:cost_fun}
\end{equation}
Finally, we can obtain the extrinsic calibration parameters $\hat{\boldsymbol{\theta}}$ and $\hat{{\bold{t}}}$ by minimizing the cost function.
\begin{equation}
    \hat{\boldsymbol{\theta}},\hat{{\bold{t}}}={\underset{\boldsymbol{\theta},\bold{t}}{\arg\min}}\mathcal{L}
    \label{eq:cost_fun_argmin}
\end{equation}

\subsection{Initialization for optimizing the cost function}
\label{sec:subsec:init}

\begin{figure}[th]
\captionsetup[subfigure]{aboveskip=-20pt,belowskip=-20pt}
\centering
\hspace{-15pt}
\subcaptionbox{\label{fig:demo_img}}{\includegraphics[width=0.25\textwidth]{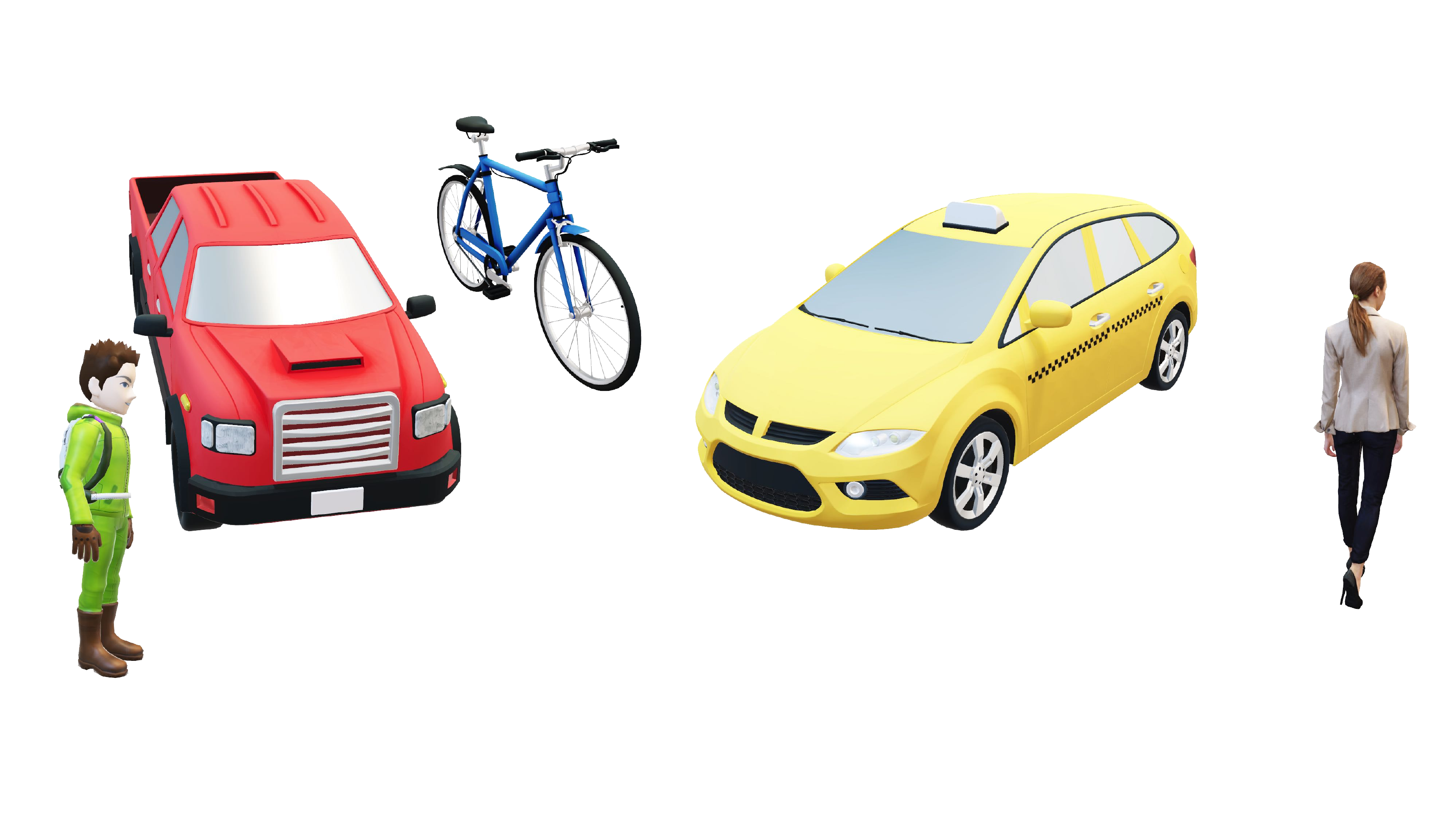}} 
\hspace{-5pt}
\subcaptionbox{\label{fig:demo_semant_img}}{\includegraphics[width=0.25\textwidth]{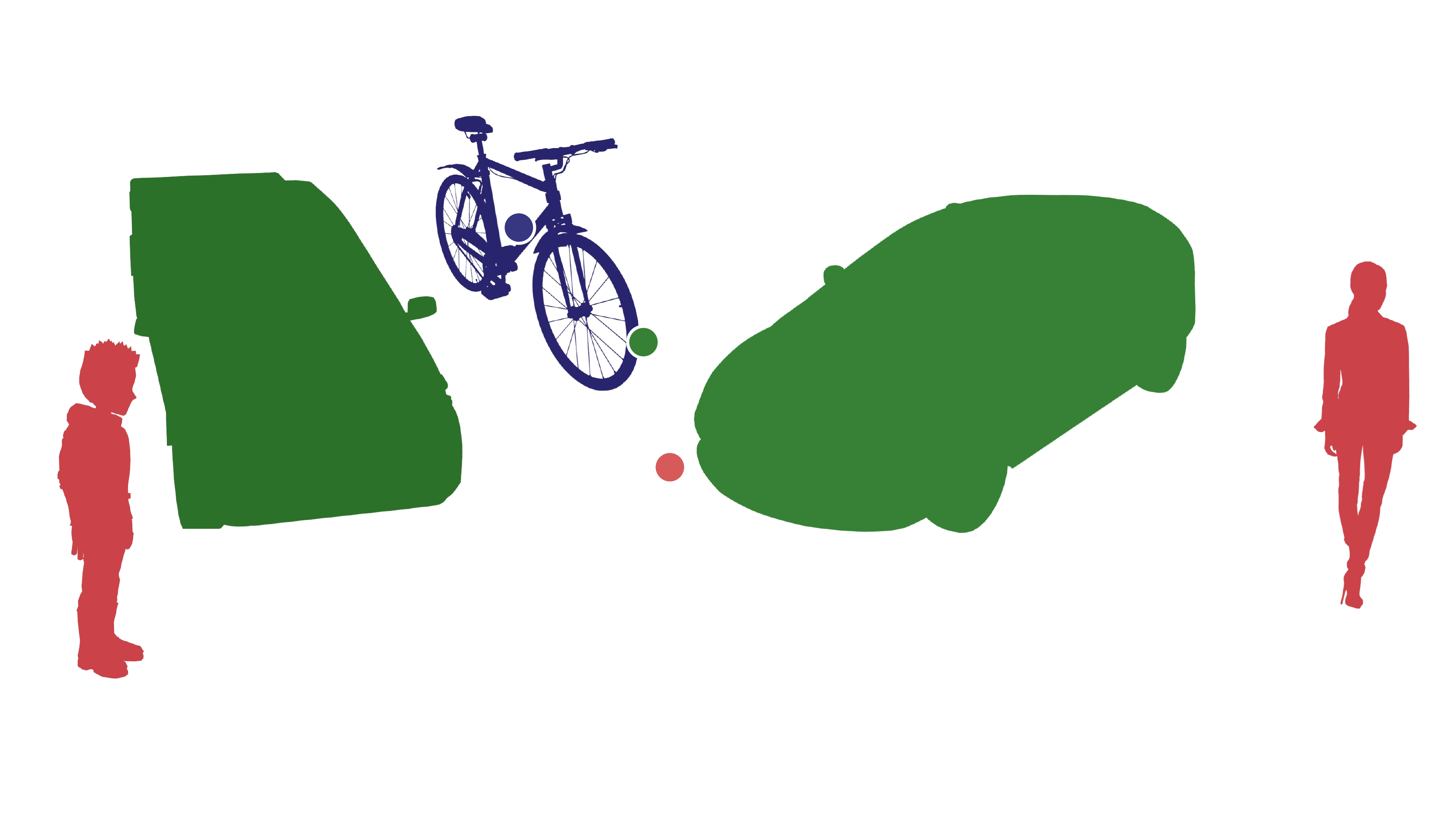}}\\
\vspace{-5pt}
\hspace{-15pt}
\subcaptionbox{\label{fig:demo_pcd}}{\includegraphics[width=0.25\textwidth]{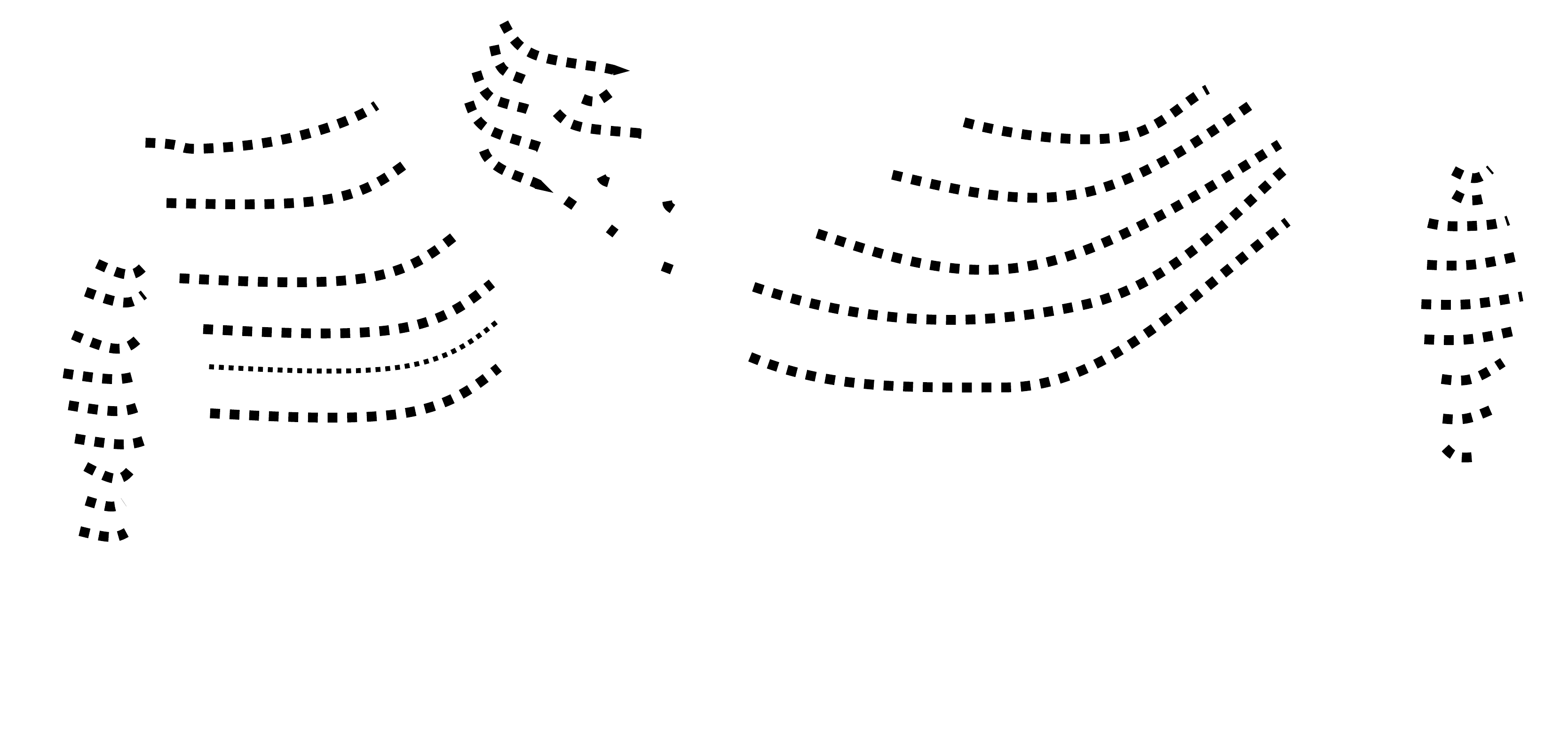}} 
\hspace{-5pt}
\subcaptionbox{\label{fig:demo_semant_pcd}}{\includegraphics[width=0.25\textwidth]{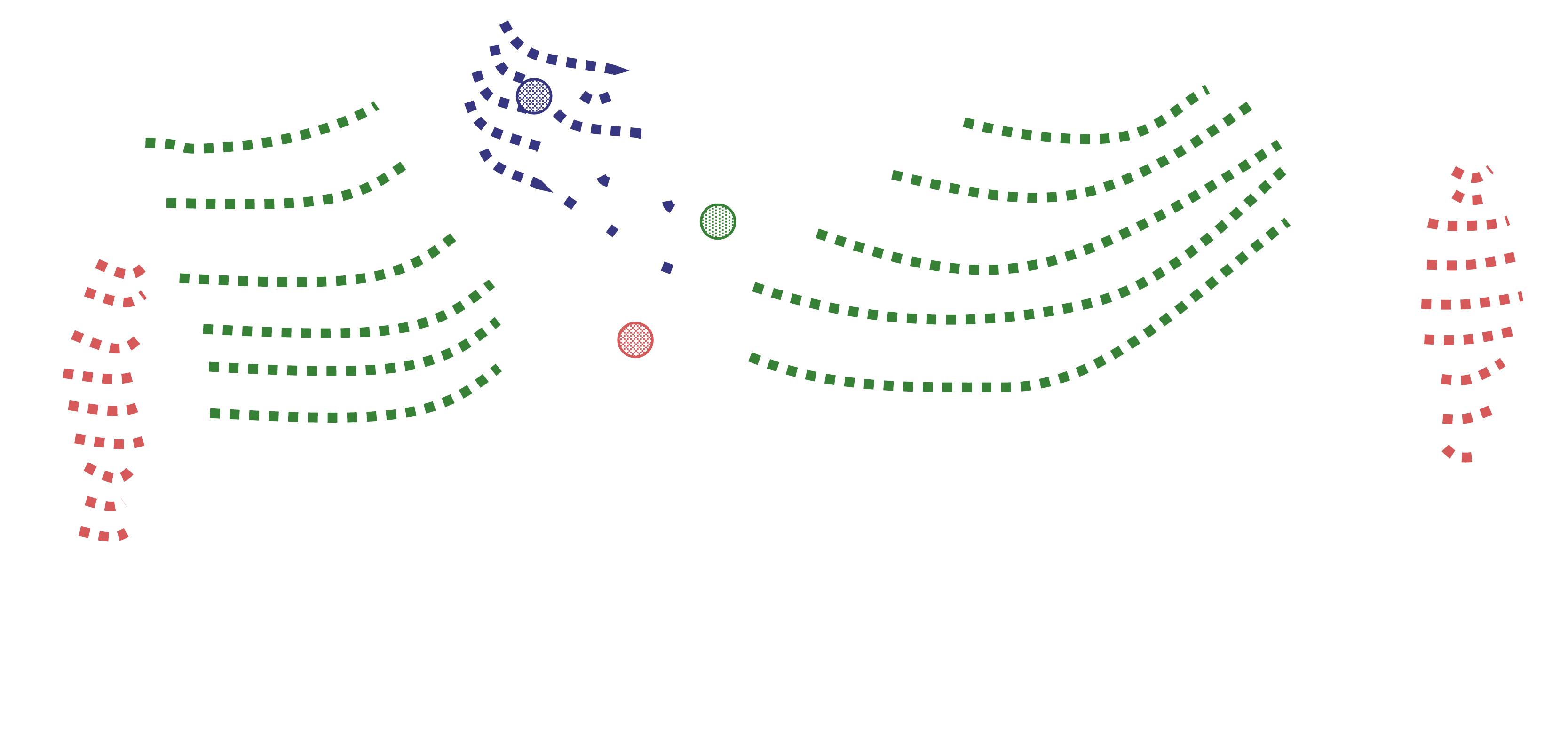}}
  \caption{(a) and (c) are RGB image and the corresponding 3D point cloud acquired by camera and LiDAR sensor. (b) and (d) are semantic segmentation results of (a) and (c). {\textcolor[rgb]{0.8,.26,.29}{Red}}, {\textcolor[rgb]{0.29, 0.52, 0.25}{green}}, {\textcolor[rgb]{0.16, 0.14, 0.47}{blue}} represents pedestrians, vehicles, bicycles class respectively. Three filled circles in (d) and (d) indicates semantic centroids (SC) of each class. }
  \label{fig:SC_demo}
\end{figure}

As stated aforehand, existing online calibration methods usually assume that rough values are known for the initialization of the optimization. This means that a preparatory target-based calibration or measurement has to be made. Inspired by the control point decision for efficiently solving the PnP problem in  EPnP \cite{Lepetit2008}, we propose the semantic centroid to form a noisy PnP problem which can be analytically solved.

As shown in Fig.~\ref{fig:SC_demo}, for the semantic centroid of point cloud with class $s$, we define the set of points with label $s$ as $P^L_s=\{ \textbf{p}_{s,1}^L,\textbf{p}_{s,2}^L,\ldots \vert  \ \textbf{p}_{s,i}^L \in P^L, \ell_i^{pcd}=s \}$ and the semantic centroid as  $\textbf{SC}_\textbf{s}^L=\frac{\sum_{\textbf{p}_i \in \{P^L_s\}}\textbf{p}_i }{\vert\{P^L_s\}\vert} $.
A similar definition can be made for the semantic centroid of images.
We consider the 3D SC of point cloud and the 2D SC of its corresponding image as a matched 3D-2D pair for the PnP problem. Note that 3D semantic centroid of the point cloud is usually not geometrically consistent with that of the corresponding image. That means 3D SCs and their corresponding 2D SCs are not well ``corresponded''. The accuracy of the extrinsic calibration parameters would not be improved even after a non-linear refinement which is ususally done for PnP problems. Thus we proposed a new cost function in Eq.~\ref{eq:cost_fun} and the derived result in this step is only used as the initial guess for optimizing the cost function.

\section{Experimental Results}
We evaluate the performance of the proposed calibration algorithm on KITTI dataset \cite{Geiger2013IJRR} and compare it with a baseline online calibration method.

\subsection{Dataset Preparation}
To investigate the effect of the quality of semantic segmentation results to the performance of the proposed method, we choose images and point clouds that have GT semantic labels as the experimental data. For images, 200 images are picked from KITTI dataset and labeled by \cite{Alhaija2018IJCV}. The classes are conformed with the Cityscapes Dataset \cite{Cordts2016Cityscapes}.
For the point cloud, we utilize the dataset released in \cite{Wu2017SqueezeSegCN}. The labels of point clouds are derived by the GT 3D bounding box provided in KITTI dataset. Three valid classes are: Vehicles, Pedestrians and Cyclists. Finally, we obtain 120 valid pairs of images who have the corresponding annotation-available point clouds. We take the extrinsic calibration parameters provided in KITTI dataset as the GT for accuracy evaluation. These parameters are derived with the offline calibration method proposed in \cite{Geiger_2012}.

\begin{figure}[t!]
\centering
    \subcaptionbox{\label{loss_fig_rot_ang}}{\includegraphics[width=0.4\textwidth]{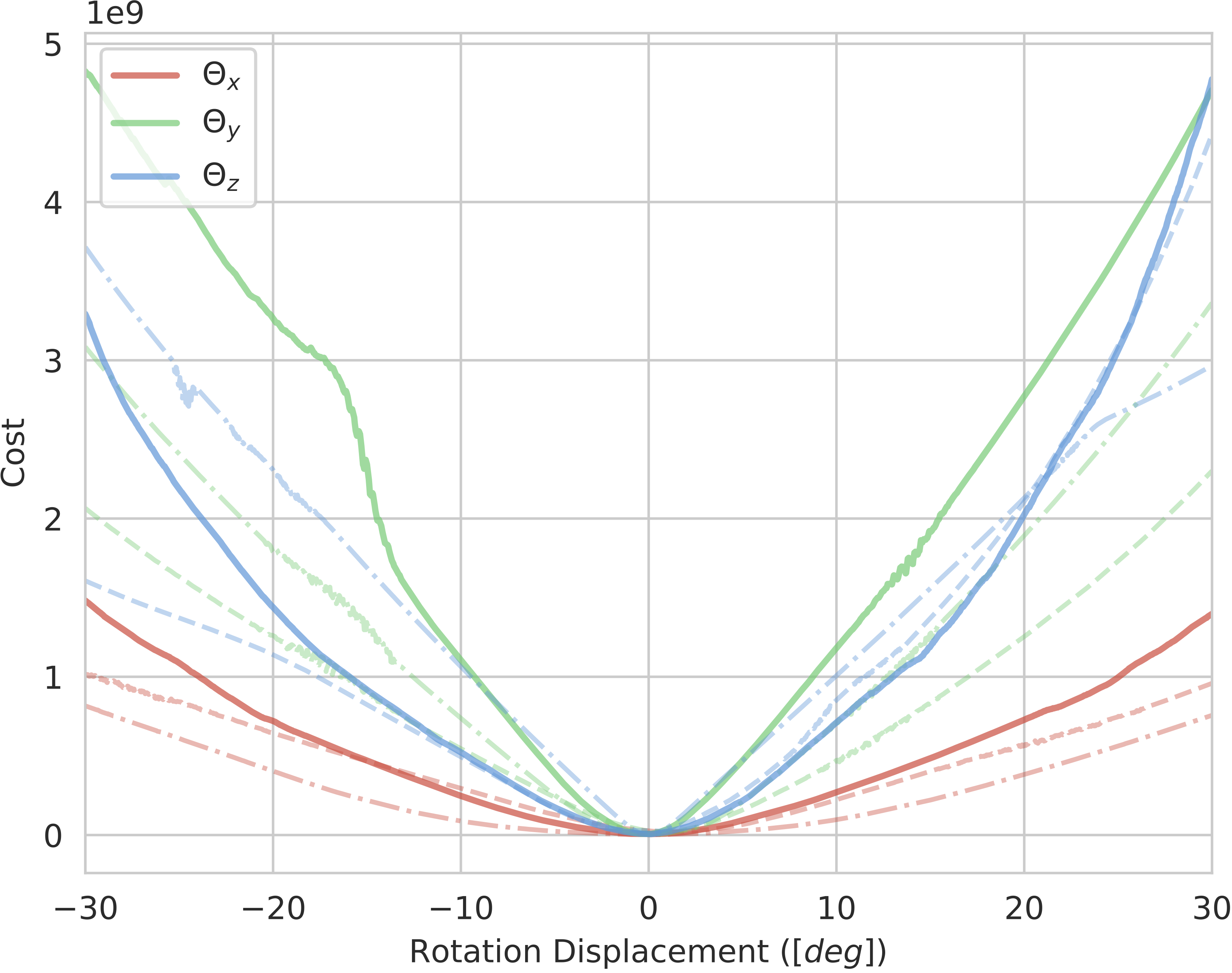}} 
    \subcaptionbox{\label{loss_fig_trans}}{\includegraphics[width=0.4\textwidth]{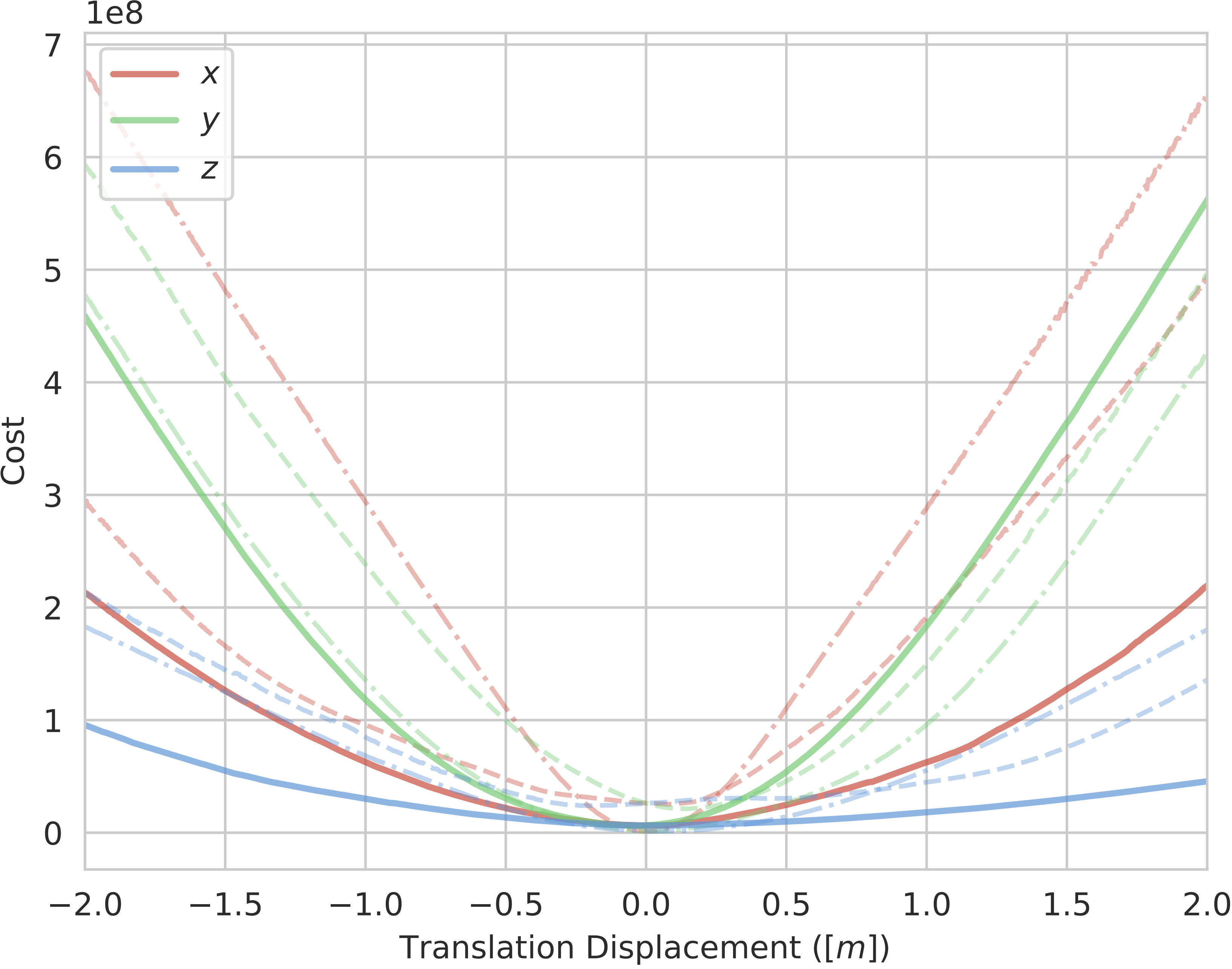}}
  \caption{Cost change of the designed cost function along with the (\protect\subref{loss_fig_rot_ang}) angular and (\protect\subref{loss_fig_trans}) translation displacement of 
  \textcolor[HTML]{e74c3c}
  {$\textbf{x}-$}, 
  \textcolor[HTML]{2ecc71}
  {$\textbf{y}-$}, 
  \textcolor[HTML]{3498db}
  {$\textbf{z}-$}axis respectively. The cost is calculated with 20 pairs. solid line
  \protect\tikz[baseline=-0.5ex]{\protect\draw [thick] (0,0) -- (0.5,0);}
  : Vehicles; dash-dot line 
  \protect\tikz[baseline=-0.5ex]{\protect\draw[thick,dash dot] (0,0) -- (0.5,0);}
  : Pedestrians; 
  dashed line 
  \protect\tikz[baseline=-0.5ex]{\protect\draw [thick,dashed] (0,0) -- (0.5,0);}
  : Cyclists. The interval for angle displacement is 0.01$^{\circ}$ and 5[mm] for translation displacement. }
  \label{loss_fig}
\end{figure}

\subsection{Convexity of the proposed cost function}
To confirm the global convergence ability of the defined cost function in Eq.~\ref{eq:cost_fun}, we plot the cost change of the cost function as the displacement occurs for each parameter in Fig.~\ref{loss_fig}. Each of the three semantic classes is calculated independently. The convexity can be confirmed in the displacement range in $ 30\degree$ with $0.01\degree$ interval for rotation parameters and $2 $[m] with $5[mm]$ interval for translation parameters. From Fig.~\ref{loss_fig_trans}, we can find that the minima of the cost plot for Vehicles class are closer to zero compared with Pedestrians and Cyclists classes. This may be caused by that vehicle's class occupied more numbers of points and pixels. Thus, we choose Vehicles class for the subsequent processing although SOIC is capable of dealing with multiple classes.

\begin{figure}[h]
\centering
    \subcaptionbox{\label{fig:vehicle_pcd_top}}{\includegraphics[width=0.4\textwidth]{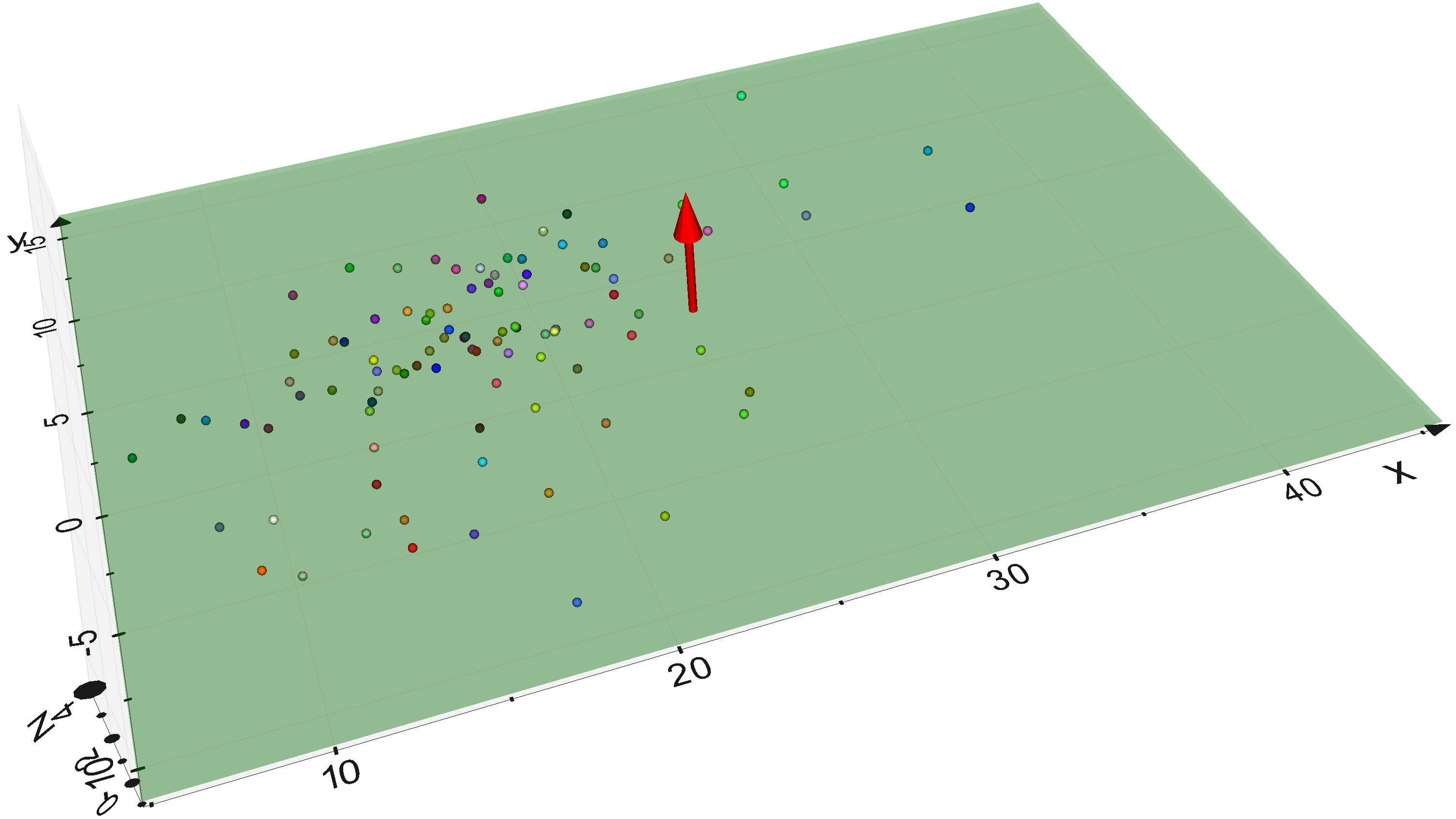}} 
    \subcaptionbox{\label{fig:vehicle_pcd_side}}{\includegraphics[width=0.5\textwidth]{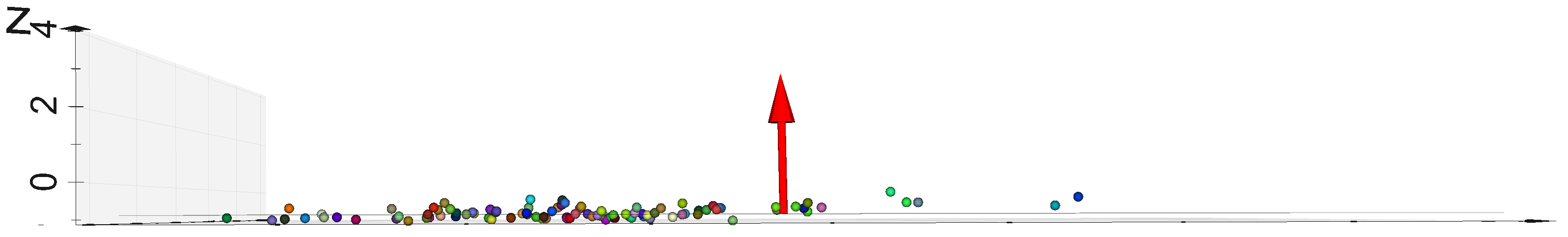}}
  \caption{Distribution of vehicles' semantic centroids of vehicles class from 100 frames.  (\protect\subref{fig:vehicle_pcd_top}) top view, (\protect\subref{fig:vehicle_pcd_side}) side view. The green plane is estimated from 3D SCs with RANSAC algorithm. The red arrows indicate the normal of the estimated plane. We can see that all centroids are approximately distributed on the same plane. }
  \label{vehicle_SCs}
\end{figure}
\begin{figure}[h!]
\centering
\includegraphics[width=0.45\textwidth]{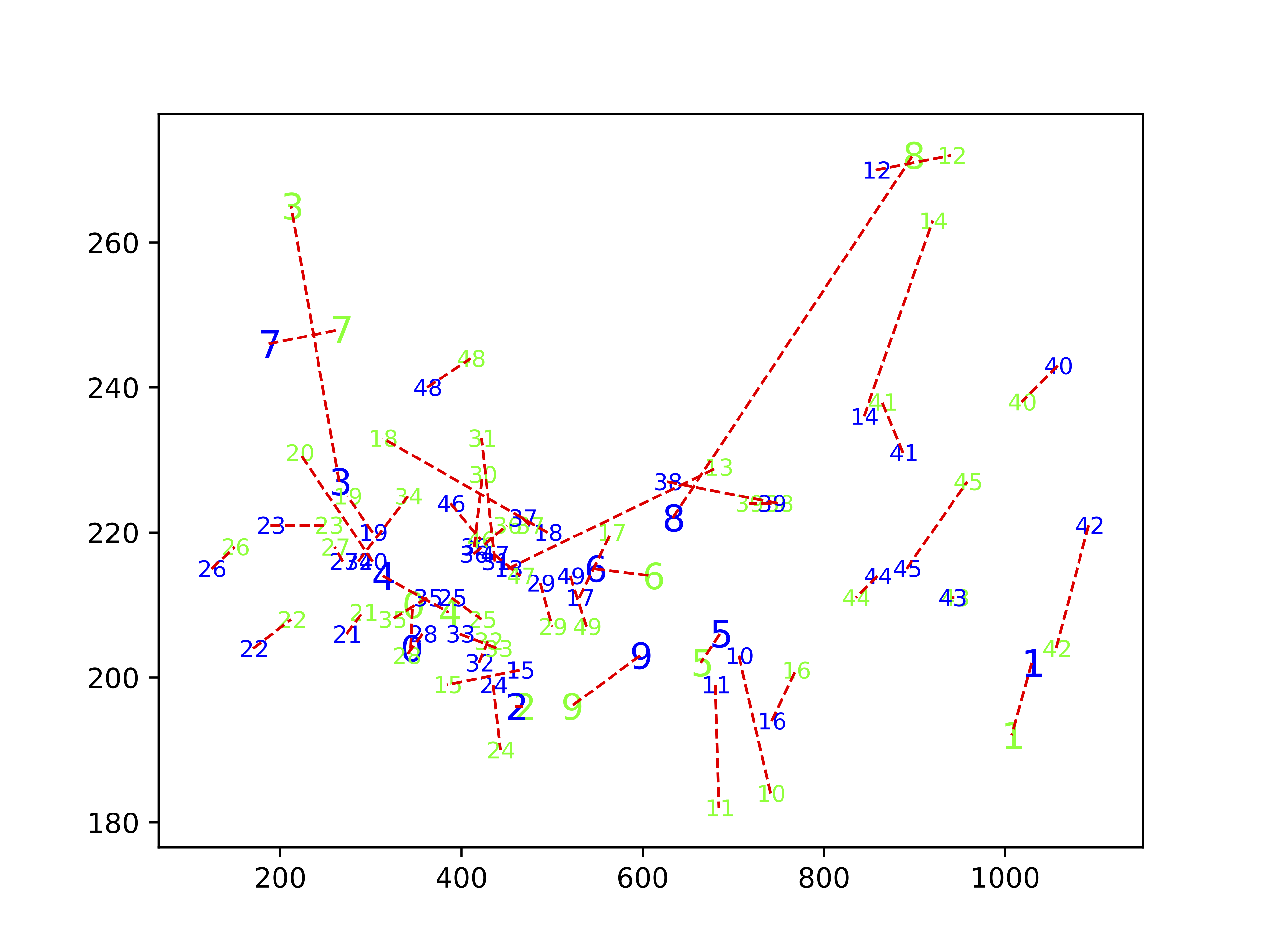}
  \caption{Correspondence of semantic centroids with the estimated initial parameters from 50 pairs. {\color{green}Green} numbers indicate semantic centroids from images and {\color{blue}blue} numbers show projected point cloud semantic centroids. The number indicate the index of the image-pointcloud pair.}
  \label{fig:proj_scs}
\end{figure}

\subsection{Initialization and optimization}
\begin{figure*}[h!]
\centering
    \mbox{
    \hspace{-20pt}
    \subcaptionbox{\label{fig:soic_pred_res:theta_x}}{\includegraphics[width=0.17\textwidth]{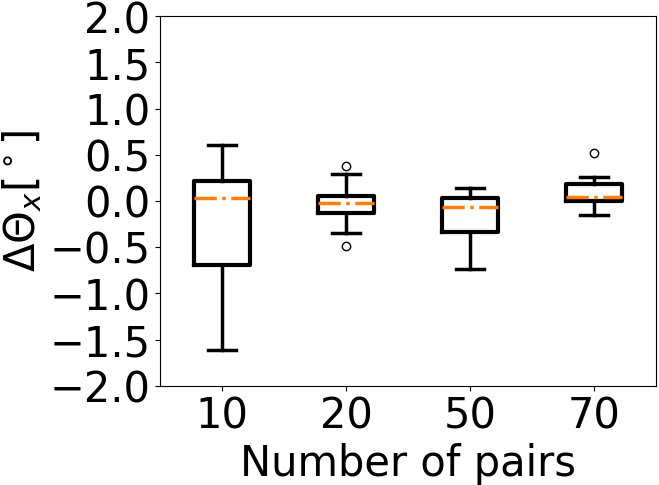}} 
    \subcaptionbox{\label{fig:soic_pred_res:theta_y}}{\includegraphics[width=0.17\textwidth]{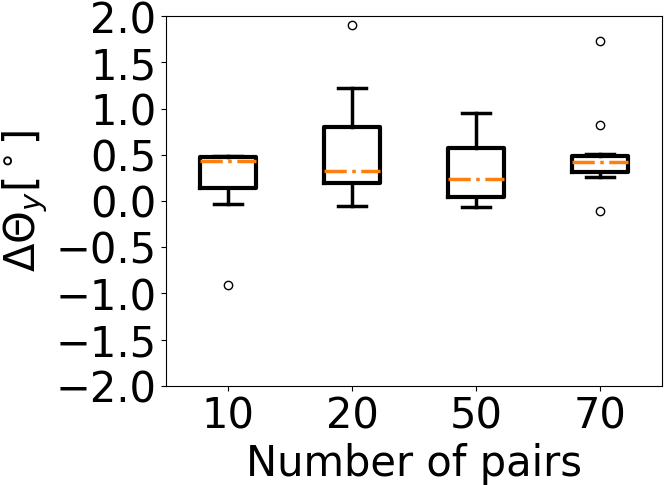}}
    \subcaptionbox{\label{fig:soic_pred_res:theta_z}}{\includegraphics[width=0.17\textwidth]{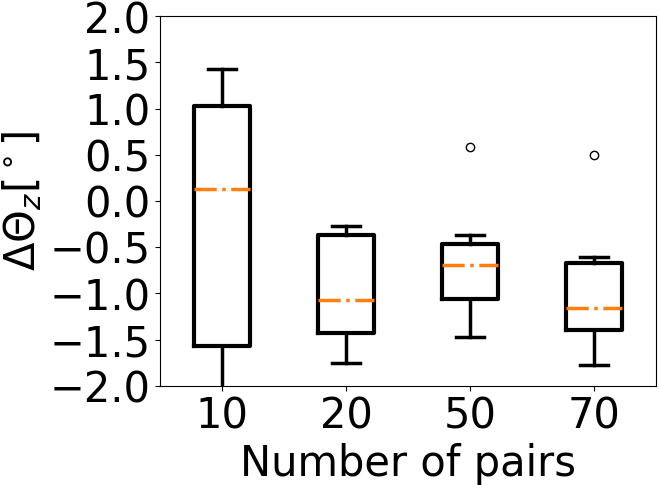}} 
    \subcaptionbox{\label{fig:soic_pred_res:x}}{\includegraphics[width=0.17\textwidth]{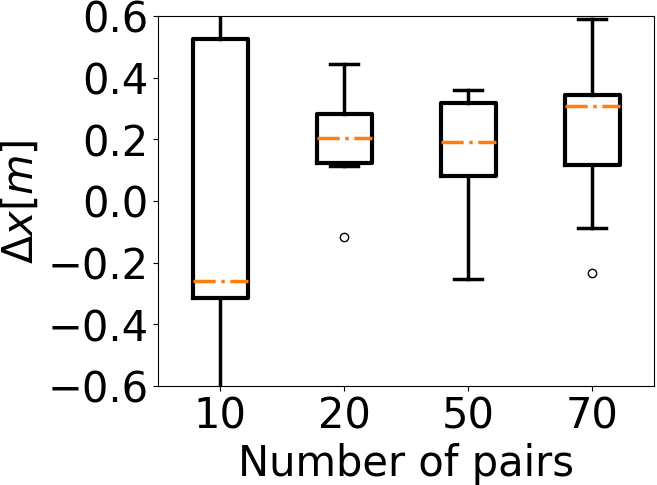}} 
    \subcaptionbox{\label{fig:soic_pred_res:y}}{\includegraphics[width=0.17\textwidth]{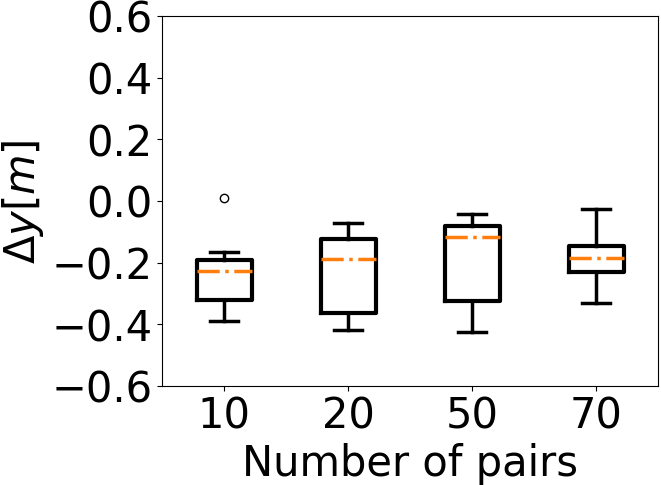}}
    \subcaptionbox{\label{fig:soic_pred_res:z}}{\includegraphics[width=0.17\textwidth]{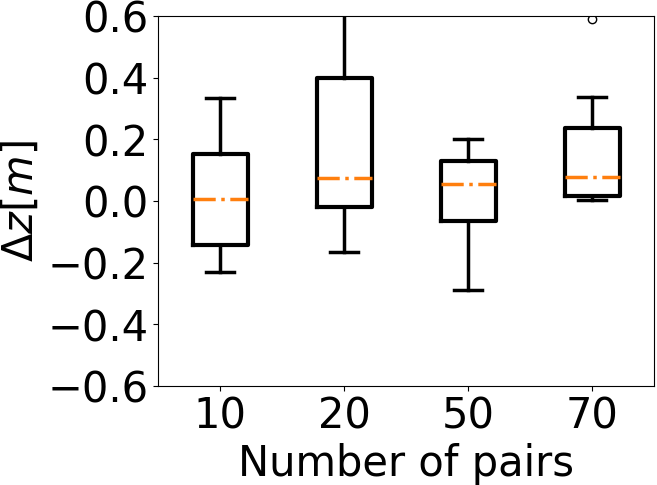}} 
    }

    \mbox{
    \hspace{-20pt}
    \subcaptionbox{\label{fig:soic_GT_res:theta_x}}{\includegraphics[width=0.17\textwidth]{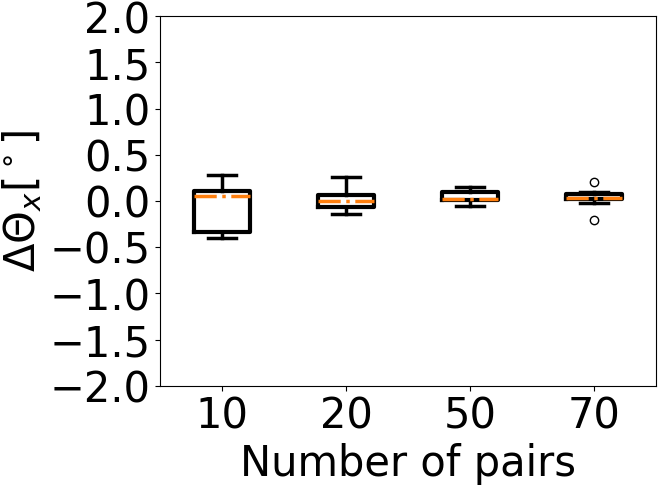}} 
    \subcaptionbox{\label{fig:soic_GT_res:theta_y}}{\includegraphics[width=0.17\textwidth]{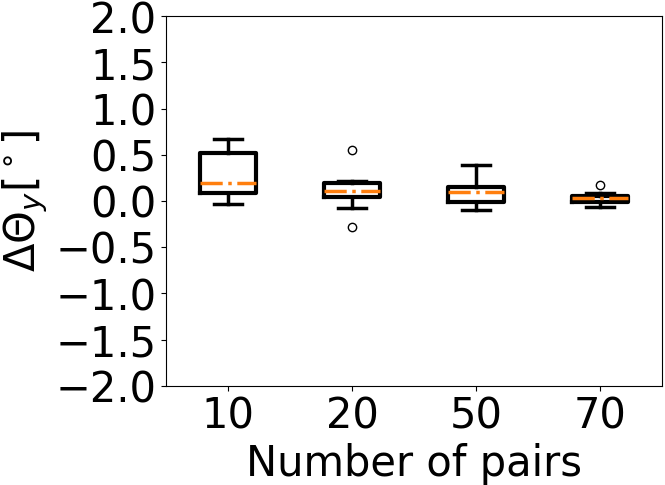}}
    \subcaptionbox{\label{fig:soic_GT_res:theta_z}}{\includegraphics[width=0.17\textwidth]{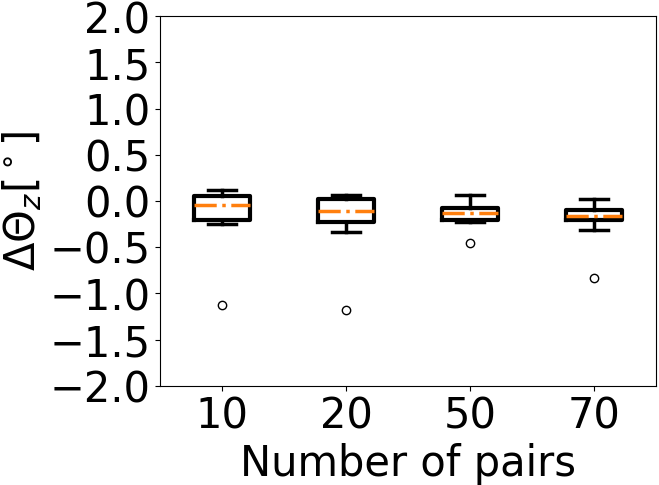}} 
    \subcaptionbox{\label{fig:soic_GT_res:x}}{\includegraphics[width=0.17\textwidth]{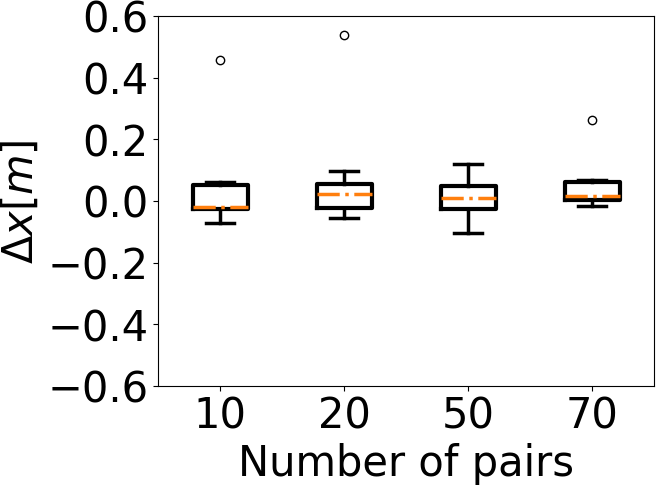}} 
    \subcaptionbox{\label{fig:soic_GT_res:y}}{\includegraphics[width=0.17\textwidth]{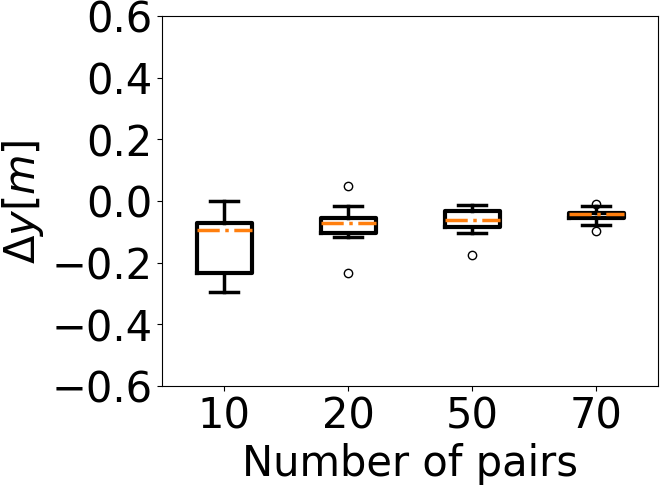}}
    \subcaptionbox{\label{fig:soic_GT_res:z}}{\includegraphics[width=0.17\textwidth]{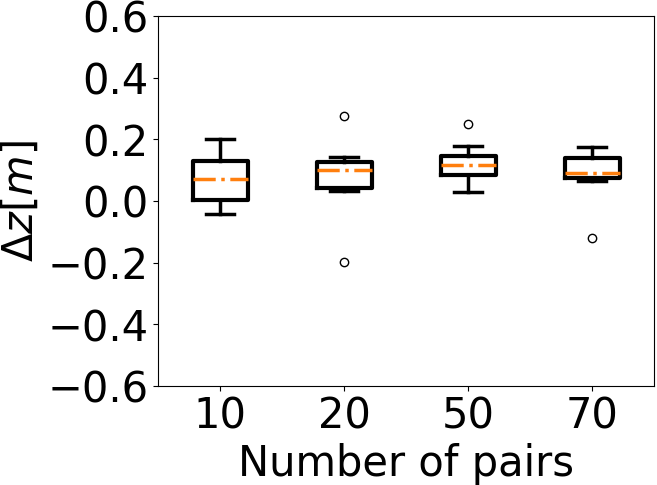}} }
  \caption{Calibration results by SOIC. 
(a)-(f): estimated results for each parameter based on predicted semantics by PointRCNN \cite{Shi_2019_CVPR} for point clouds and NVIDIA semantic segmentation model \cite{semantic_cvpr19} for images; (g)-(l) results with GT semantics. For each number of pairs, we perform SOIC 10 times on randomly selected point cloud and image pairs.}
  \label{fig:soic_res}
\end{figure*}

\begin{figure*}[h!]
\centering
    \mbox{
    \hspace{-20pt}
    \subcaptionbox{\label{fig:MI_GT_init:theta_x}}{\includegraphics[width=0.17\textwidth]{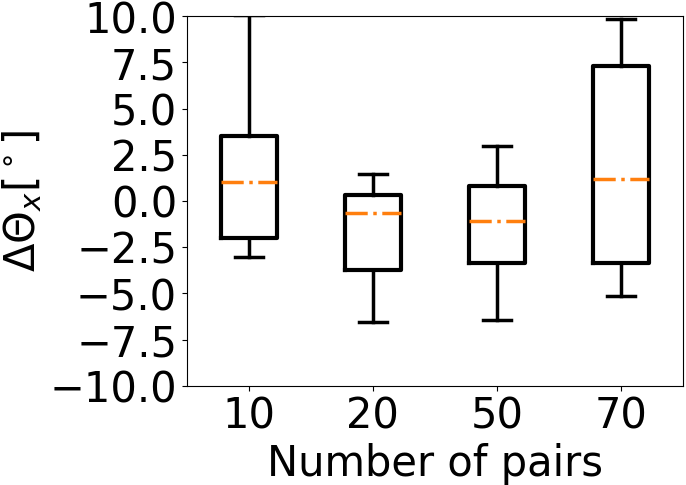}} 
    \subcaptionbox{\label{fig:MI_GT_init:theta_y}}{\includegraphics[width=0.17\textwidth]{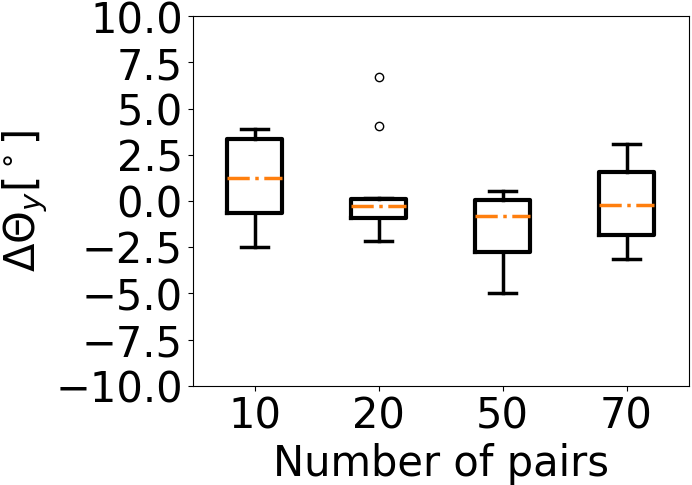}}
    \subcaptionbox{\label{fig:MI_GT_init:theta_z}}{\includegraphics[width=0.17\textwidth]{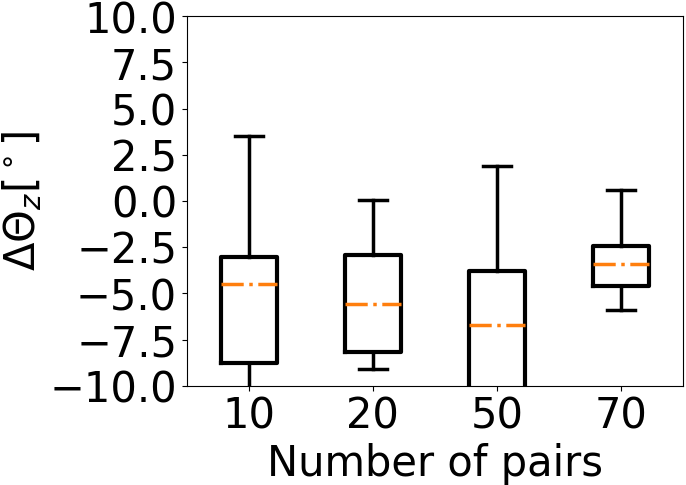}}
    \subcaptionbox{\label{fig:MI_GT_init:x}}{\includegraphics[width=0.17\textwidth]{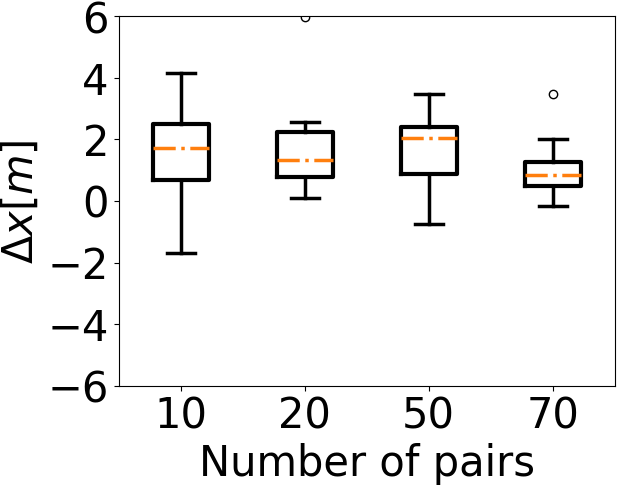}} 
    \subcaptionbox{\label{fig:MI_GT_init:y}}{\includegraphics[width=0.17\textwidth]{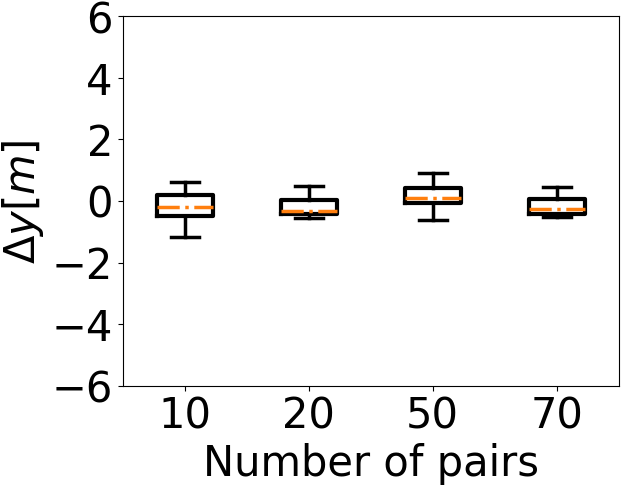}}
    \subcaptionbox{\label{fig:MI_GT_init:z}}{\includegraphics[width=0.17\textwidth]{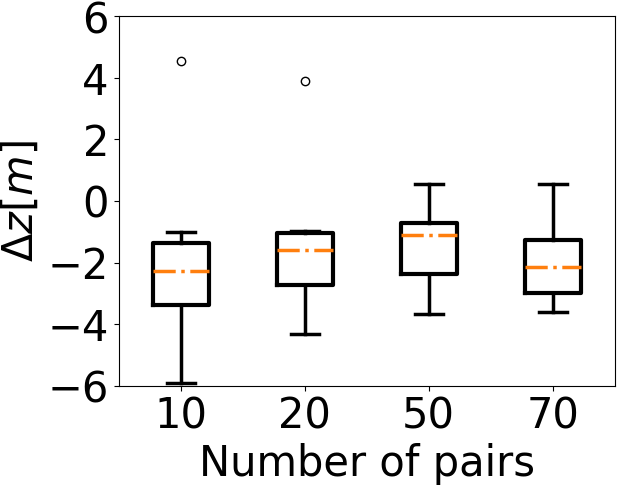}} }

    \mbox{
    \hspace{-20pt}
    \subcaptionbox{\label{fig:MI_SOIC_GT_init:theta_x}}{\includegraphics[width=0.17\textwidth]{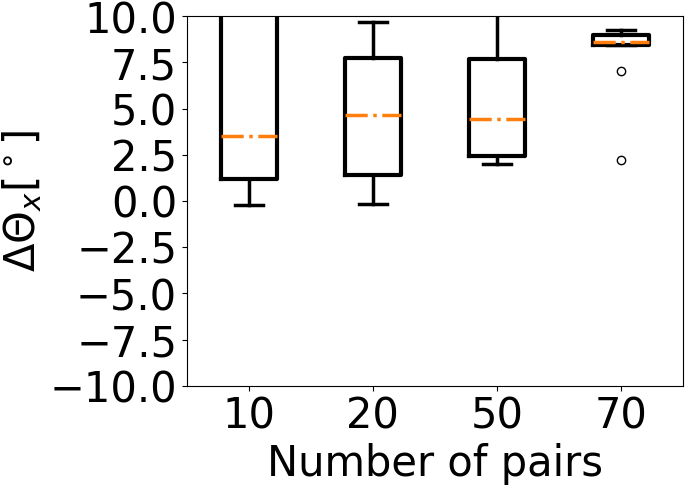}} 
    \subcaptionbox{\label{fig:MI_SOIC_GT_init:theta_y}}{\includegraphics[width=0.17\textwidth]{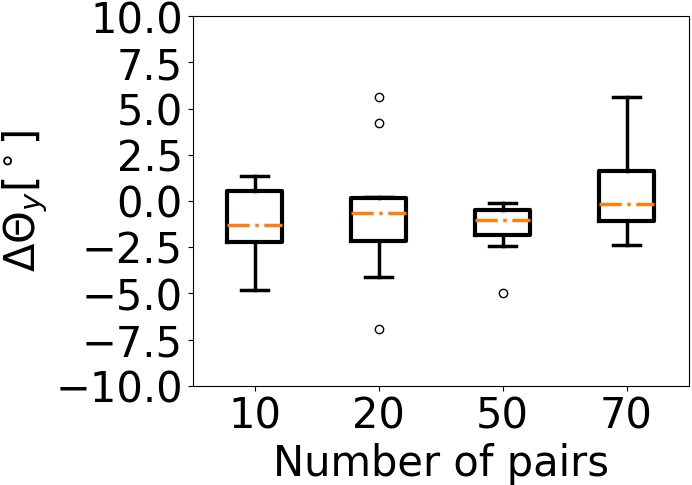}}
    \subcaptionbox{\label{fig:MI_SOIC_GT_init:theta_z}}{\includegraphics[width=0.17\textwidth]{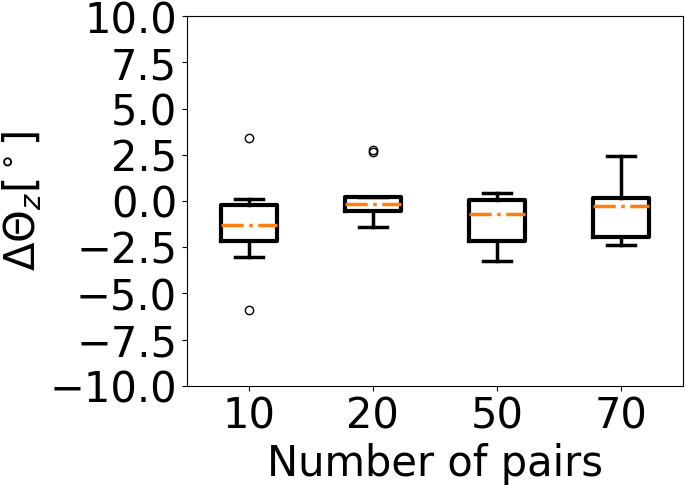}} 
    \subcaptionbox{\label{fig:MI_SOIC_GT_init:x}}{\includegraphics[width=0.17\textwidth]{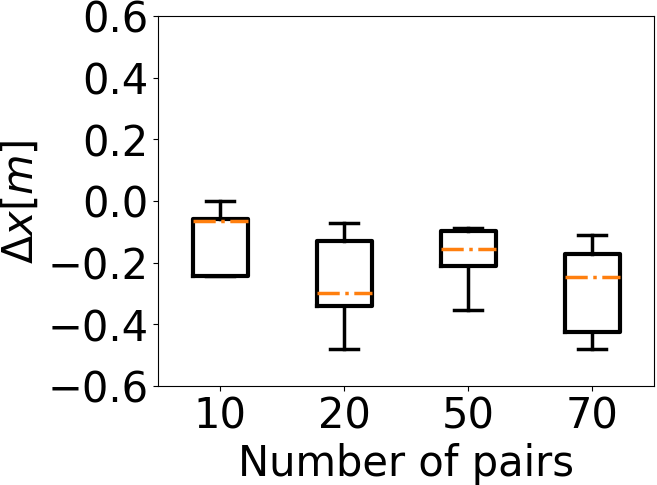}} 
    \subcaptionbox{\label{fig:MI_SOIC_GT_init:y}}{\includegraphics[width=0.17\textwidth]{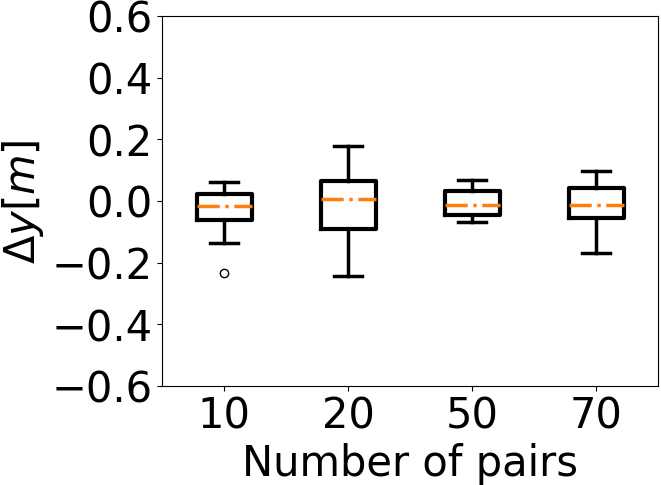}}
    \subcaptionbox{\label{fig:MI_SOIC_GT_init:z}}{\includegraphics[width=0.17\textwidth]{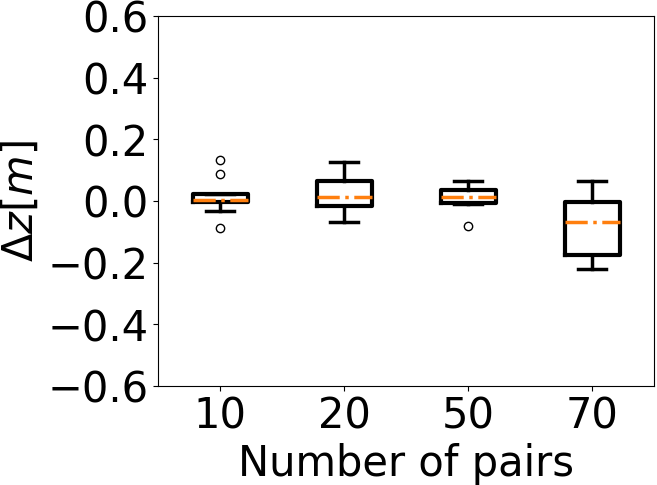}}}
  \caption{Calibration results by MI \cite{MIcali}. (a)-(f): Results by MI with the same initial guesses estimated by SOIC; (g)-(l): Results with GT extrinsic values as the initial guess. Note that the range of y axis in (a)-(c) and (g)-(i) is 5 times, in (d)-(f) and (j)-(l) is 10 times greater than that in Fig.\ref{fig:soic_res}.}
  \label{fig:MI_res}
\end{figure*}

As defined in Sec. \ref{sec:subsec:init}, we can obtain the 3D-2D SC pairs for PnP problems with Vehicles class. As shown in Fig.~\ref{vehicle_SCs}, we interestingly find that all SCs distributed approximately on a plane after visualizing 3D semantic centroids of 100 point clouds. This makes it difficult to solve the PnP problem with usual methods such as EPnP \cite{Lepetit2008}. With this observation, we adopt the Infinitesimal Plane-based Pose Estimation (IPPE) algorithm proposed in \cite{Collins2014InfinitesimalPP} to derive the initial extrinsic values.

With the estimated initial parameters, we project 3D semantic centroids of point clouds to the image plane as Fig.~\ref{fig:proj_scs} shows. As describe in \S\ref{sec:subsec:init}, the correspondences between 2D-3D SCs are erroneous in most cases. Even so, we can see that 3D SCs are roughly projected to pixels close to their corresponding SCs.

We use Powell's conjugate direction method \cite{Powell1964AnEM} as the optimization method to find extrinsic calibration parameters that minimize the cost function in Eq.~\ref{eq:cost_fun}.

\subsection{Semantic segmentation of point clouds and images}
The GT semantic labels can help to show the accuracy upper bound of the proposed method. On the other hand, it is impractical to predict the segmentation labels as accurate as GT in the real applications.
For practical purposes, we apply SOIC with semantics predicted by real networks. PointRCNN \cite{Shi_2019_CVPR} is used for the point cloud semantic segmentation. A pre-trained model on KITTI dataset for Vehicles class is provided. Nvidia Semantic Segmentation \cite{semantic_cvpr19} is used for image semantic segmentation. A pre-trained model on KITTI dataset with WideResNet38 backbone is provided for segmentation prediction.
Note that the point clouds and images for evaluating SOIC may be included in the training dataset for the two pre-trained models. We think this will not affect a lot as they are used to demonstrate the capability of SOIC to work with real predicted semantics. Other SOTA models can be adopted to get better semantic segmentation performance.

\subsection{Quantitative Results}
To also investigate how many pairs are needed, we apply SOIC to estimate the extrinsic calibration parameters with $10, 20, 50, 70$ pairs respectively. For each number of pairs, we perform the calibration process 10 times with randomly sampled point cloud and image pairs. For example, for 20 pairs, we take 10 groups of 20 pairs of point cloud and images which are randomly sampled. Then, we perform SOIC on each of these 10 groups.
The calibration results for each parameter are listed in Fig.~\ref{fig:soic_res}. Results with semantics predicted by pre-trained models are from Fig.\ref{fig:soic_res} (a)-(f). Results with GT semantics are from Fig.\ref{fig:soic_res} (g)-(l).  

Evaluations with the same data on MI \cite{MIcali} are also performed. The results are listed in Fig.~\ref{fig:MI_res}.
Results estimated by MI with the initial values estimated by SOIC are shown from Fig.\ref{fig:MI_res} (a)-(f).  Estimated results by MI with GT parameters as the initial guess is from Fig.\ref{fig:MI_res} (g)-(l). 
Note that the range of rotation errors in Fig.~\ref{fig:soic_res} is 5 times greater than that in Fig.~\ref{fig:MI_res}. And translation errors in Fig.~\ref{fig:soic_res} (d)-(f) is 10 times greater than that in Fig.~\ref{fig:MI_res}. By comparing (a)-(f) in Fig.~\ref{fig:soic_res} and Fig.~\ref{fig:MI_res}, we can find the SOIC is more robust to the erroneous initial guess. 

To show the errors more visually, we additionally pick up the best case for each number of pairs and show the errors in Tab. \ref{tab:cali_res}.  For SOIC, the calibrated parameters with the least cost are selected from 10 trials. For MI, we manually selected parameters with the least errors. From this table, we consider 20 pairs can generate calibration results with errors approximately $\pm1\degree$ for rotation and $\pm0.1$[m] for translation with the semantic segmentation result predicted by pre-trained models in this work.

\newcolumntype{P}[1]{>{\centering\arraybackslash}p{#1}}
\newcolumntype{M}[1]{>{\centering\arraybackslash}m{#1}}

\begin{table}[t]
\centering
\resizebox{.5\textwidth}{!}{%
\hspace{-10pt}
\begin{tabular}{|c|c|M{1.2cm}|M{1.2cm}|M{1.2cm}|M{1.2cm}|}
\hline
\multicolumn{2}{|l|}{\diagbox[innerwidth=2.5cm]{\\\hspace{-.22cm}\textbf{\small Parameters}\\\hspace{-.12cm} \textbf{\footnotesize estimated} \\\hspace{-.3cm} \textbf{\footnotesize{with different pairs}}}{\\ \textbf{\small Method}}} & \begin{tabular}[c]{@{}c@{}}\hspace{-.2cm}MI \cite{MIcali}\\\hspace{-.18cm}(initial guess\\\hspace{-.2cm}by SOIC\\\hspace{-.1cm} from GT\\semantics)\end{tabular} & \begin{tabular}[c]{@{}c@{}}\hspace{-.2cm}MI \cite{MIcali}\\\hspace{-.18cm}(initial guess\\\hspace{-.2cm}with GT\\\hspace{-.1cm} calibration\\ values)\end{tabular} & \begin{tabular}[c]{@{}c@{}}SOIC\\ (w/ Pred. \\ semantics)\end{tabular} & \begin{tabular}[c]{@{}c@{}}SOIC\\ (w/ GT \\ semantics)\end{tabular} \\ \hline
\multirow{4}{*}{$\Delta\Theta_x[\degree]$} & 10 &\ApG{-1.395} &\ApG{-0.237}  & \ApG{-0.504} & \ApG{-0.399} \\ \cline{2-5} 
 & 20 &\ApG{-1.149} &\ApG{-0.144}   & \ApG{-0.352} & \ApG{-0.075} \\ \cline{2-5} 
 & 50 &\ \ApG{0.775}  &\ \ApG{2.455}  & \ \ApG{0.042} & \ \ApG{0.070} \\ \cline{2-5} 
 & 70 &\ApG{-1.547} & \ \ApG{2.233} & \ApG{-0.053} & \ \ApG{0.016} \\ \hline\hline
\multirow{4}{*}{$\Delta\Theta_y[\degree]$} & 10 &\ApG{-0.930}  & \ \ApG{0.530} &\ApG{0.302} &\ 0.090 \\ 
\cline{2-5} 
 & 20 &\ \ApG{0.006} & \ApG{-0.036} & \ApG{-0.059} & \ApG{-0.073}\\ \cline{2-5} 
 & 50 &\ \ApG{0.533} & \ApG{-0.563}  &\ \ \ApG{0.037}  &\  \ApG{0.171} \\ \cline{2-5} 
 & 70 & \ApG{-1.330}& \ApG{-1.390}  &\ \ApG{0.331} & -0.001 \\ \hline\hline
\multirow{4}{*}{$\Delta\Theta_z[\degree]$} & 10 & \ \ApG{3.520}& \ \ApG{0.119} &\ \ \ApG{0.613} &\ 0.051 \\ \cline{2-5} 
 & 20 &\ \ApG{0.075} &  \ \ApG{0.023} &  \ApG{-0.447} & \ApG{-0.177} \\ \cline{2-5} 
 & 50 & \ApG{-1.179}& \ApG{-0.533} & \ApG{-0.894} & \ApG{-0.233} \\ \cline{2-5} 
 & 70 &\ \ApG{0.572} & \ApG{-0.570} & \ApG{-0.605} & \ApG{-0.132} \\ \hline\hline
\multirow{4}{*}{$\Delta t_x$[m]} & 10 &\ApG{-1.693} &  \ApG{-0.065} &  \ApG{-0.291} & -0.047 \\ \cline{2-5} 
 & 20 &\ \ApG{0.274} & \ApG{-0.269} &\ \ApG{0.115} &\ \ApG{0.041} \\ \cline{2-5} 
 & 50 & \ \ApG{0.500}&\ApG{-0.097}  &\ \ApG{0.168} &\ \ApG{0.061} \\ \cline{2-5} 
 & 70 & \ \ApG{0.427}& \ApG{-0.110} &\ \ApG{0.185} &\ \ApG{0.049} \\ \hline\hline
\multirow{4}{*}{$\Delta t_y$[m]} & 10 & \ \ApG{0.304}&  \ApG{-0.016}& \ApG{-0.214} & -0.085 \\ \cline{2-5} 
 & 20 &\ApG{-0.555}  &  \ \ApG{0.026} & \ApG{-0.072} & \ApG{-0.015} \\ \cline{2-5} 
 & 50 &\ \ApG{0.013} & \ \ApG{0.036} & \ApG{-0.065} & \ApG{-0.086} \\ \cline{2-5} 
 & 70 & \ \ApG{0.081}& \ApG{-0.015} & \ApG{-0.172} & \ApG{-0.042} \\ \hline\hline
\multirow{4}{*}{$\Delta t_z$[m]} & 10 &  \ApG{-1.352}& \ \ApG{0.010}  &\ \ApG{0.275} &\ \ApG{0.003} \\ \cline{2-5} 
 & 20 &\ApG{-3.050} & \ApG{-0.070} &\ \ApG{0.028} &\ \ApG{0.130} \\ \cline{2-5} 
 & 50 & \ApG{-3.289}& \ApG{-0.010} &\ \ApG{0.072} &\ \ApG{0.090} \\ \cline{2-5} 
 & 70 & \ApG{-2.096} & \ \ApG{0.002} &\ \ApG{0.055} &\ \ApG{0.078} \\ \hline
\end{tabular}%
}
\caption{Errors of calibration result by MI and SOIC under different conditions. Darker color indicates the greater error. For SOIC, the calibrated parameters with the least cost are selected from 10 trials. For MI, we manually selected parameters with the least errors.}
\label{tab:cali_res}
\end{table}

\subsection{Qualitative evaluation}
\begin{figure}[htpb]
\centering
    \captionsetup[subfigure]{aboveskip=0pt,belowskip=-.4pt}
    \subcaptionbox{\label{fig:raw_img_95}}{\includegraphics[width=0.44\textwidth]{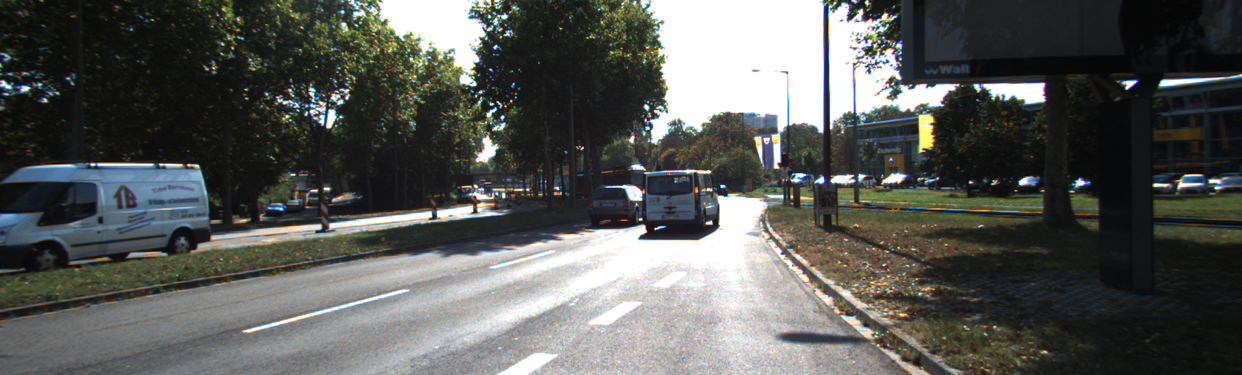}} \\
    \subcaptionbox{\label{fig:oic_pred_95}}{\includegraphics[width=0.44\textwidth]{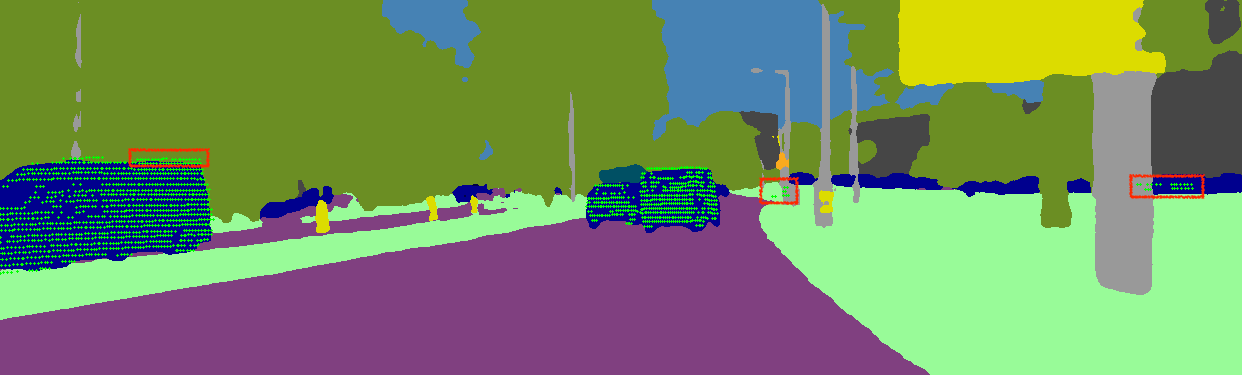}}\\
    \subcaptionbox{\label{fig:soic_GT_95}}{\includegraphics[width=0.44\textwidth]{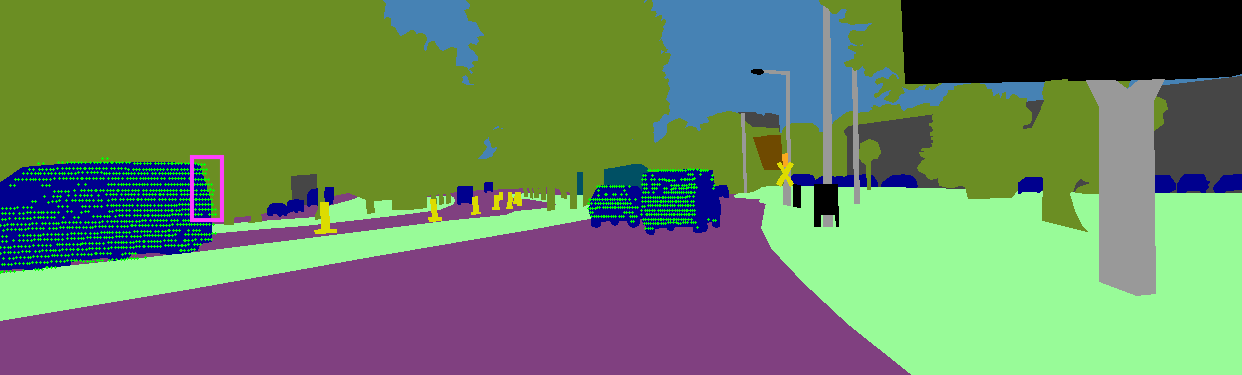}}\\
    \subcaptionbox{\label{fig:GT_GT_95}}{\includegraphics[width=0.44\textwidth]{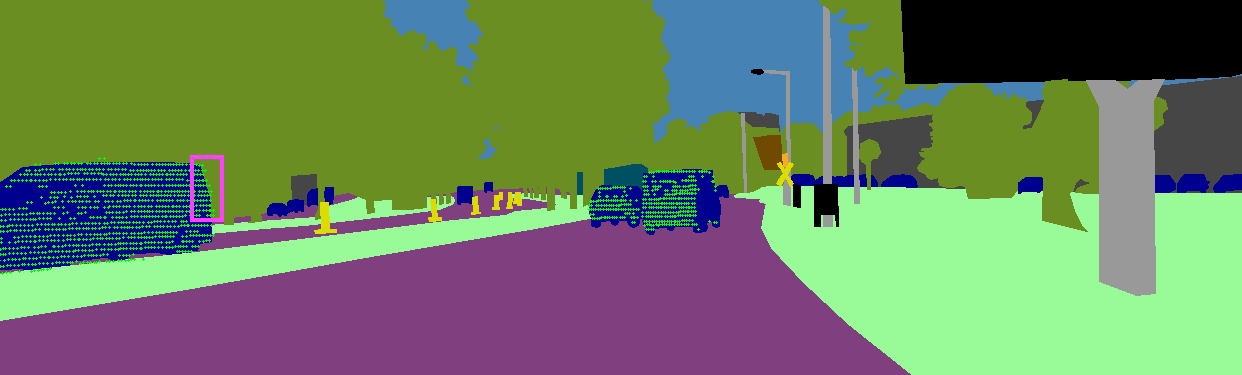}}
    \caption{An example qualitative result. (a): Raw RGB image. 
    (b): Projection of predicted semantic 3D points to the corresponding predicted semantic image with extrinsic parameters estimated by SOIC w/ predicted semantics.
    (c): Projection of GT semantic 3D points to the corresponding GT semantic image with extrinsic parameters estimated by SOIC w/ GT semantics
    (d): Projection of GT semantic 3D points to the corresponding GT semantic image with GT extrinsic parameters.
    For semantic images, the color are conformed with the Cityscapes Dataset \cite{Cordts2016Cityscapes}. Green points in (b)-(d) are the vehicle's class predicted by PointRCNN. Red boxes in (b) show the points that are segmented as Vehicles class wrongly by PointRCNN. The pink box in (c) shows the slight mis-alignment compared with that in (d).}
    \label{fig:quali_res}
\end{figure}

We pick up an example scene to show the qualitative results in Fig.~\ref{fig:quali_res}. The semantic segmentation result of image Fig.~\ref{fig:quali_res}a with NVIDIA semantic segmentation model is showed in .Fig.~\ref{fig:quali_res}b. Green points in Fig.~\ref{fig:quali_res}b shows the 3D point with vehicles class predicted by PointRCNN.  Projection with extrinsic parameters estimated by SOIC with predicted semantics, with GT semantics and GT parameters are showed in \ref{fig:quali_res}b-d respectively. By comparing with \ref{fig:quali_res}c,d, mis-segemented pixels/points predicted by pre-trained models can be found on the Fig.~\ref{fig:quali_res}(b). Slight mis-alignment caused by the calibration error is expected to be showed by comparing the content of red box in Fig.~\ref{fig:quali_res}c and Fig.~\ref{fig:quali_res}d.

\subsection{Discussions}
Although the proposed method shows higher accuracy than baseline method, there still is a gap to the GT level especially for translation parameters.  
We consider the error may come from: 1) Semantic segmentation errors. 2) Inherent mis-alignment due to movement of dynamic objects.
We believe there is much room to improve the accuracy of SOIC. For the first aspect, we can utilize models with higher segmentation performance. A semi-supervised based on the semantic matching can a be possible future work to improve the segmentation accuracy. For the second aspect, we can select static objects as semantics for SOIC.
For outdoor scenes, we can choose the image and point cloud data when the sensors are static. The drift of synchronization between point cloud data and images can be decreased. Also, the target semantic objects can be static things such as traffic signs. For indoor scenes, we can intentionally set static objects, for example chairs or stationary humans.
In general, utilizing more static semantic objects would increase accuracy. 

\section{CONCLUSION}
In this paper, we presented a novel semantic online initialization calibration method, SOIC, that can estimate the extrinsic calibration parameters between LiDAR and camera sensors. By utilizing the semantic information, SOIC owns the advantages of both no need for specific targets and no need for initial values.

On the other hand, as there is no need for the initial parameters, the proposed method can even be used for ``start-from-scratch'' scenarios. Furthermore, the proposed approach can be easily extended to calibrate any other modal sensor pairs (e.g., RGB camera-FIR camera, camera-radar) as long as semantics can be observed.

\bibliographystyle{IEEEtran}
\bibliography{ref}

\end{document}